\documentclass[10pt,twocolumn,letterpaper]{article}
\usepackage[pagenumbers]{wacv} 

\usepackage{times}
\usepackage{epsfig}
\usepackage{graphicx}
\usepackage{amsmath}
\usepackage{amssymb}
\usepackage{booktabs}
\usepackage{array,multirow}
\usepackage{xcolor}
\usepackage[nolist]{acronym}
\usepackage{bbm}
\usepackage{colortbl}
\usepackage[T1]{fontenc}

\newcommand{\minisection}[1]{\vspace{1mm}\noindent{\textbf{#1}.}}

\newenvironment{tight_itemize}{
\begin{itemize}
  \setlength{\topsep}{0pt}
  \setlength{\itemsep}{2pt}
  \setlength{\parskip}{0pt}
  \setlength{\parsep}{0pt}
}{\end{itemize}}

\usepackage{color}
\usepackage[mathscr]{euscript}

\newcommand{\bali}{\begin{aligned}}
	\newcommand{\eali}{\end{aligned}}

\newcommand{\Ebb}[0]{\ensuremath{\mathbb{E}} }

\newcommand{\onev}[0]{\ensuremath{{\bf 1}} }

\newcommand{\argmin}{\operatornamewithlimits{argmin}}

\usepackage[pagebackref=true,breaklinks=true,colorlinks,bookmarks=false]{hyperref}

\usepackage[capitalize]{cleveref}
\crefname{section}{Sec.}{Secs.}
\Crefname{section}{Section}{Sections}
\Crefname{table}{Table}{Tables}
\crefname{table}{Tab.}{Tabs.}

\begin{document}

\title{HyperMix: Out-of-Distribution Detection and Classification in Few-Shot Settings}

\author{
  Nikhil Mehta$^{1, 2, *}$\quad Kevin J Liang$^{2, *}$\quad Jing Huang$^2$\quad Fu-Jen Chu$^2$\quad Li Yin$^2$\quad Tal Hassner$^2$ \\
  Duke University$^1$ \quad Meta$^2$ \quad Equal Contribution$^*$\\
  \texttt{nikhilmehta.dce@gmail.com} \\
}
\maketitle

\begin{abstract}
   Out-of-distribution (OOD) detection is an important topic for real-world machine learning systems, but settings with limited in-distribution samples have been underexplored.
  Such few-shot OOD settings are challenging, as models have scarce opportunities to learn the data distribution before being tasked with identifying OOD samples.
  Indeed, we demonstrate that recent state-of-the-art OOD methods fail to outperform simple baselines in the few-shot setting.
  We thus propose a hypernetwork framework called HyperMix, using Mixup on the generated classifier parameters, as well as a natural out-of-episode outlier exposure technique that does not require an additional outlier dataset.
  We conduct experiments on CIFAR-FS and MiniImageNet, significantly outperforming other OOD methods in the few-shot regime.
\end{abstract}

\section{Introduction}

\ac{OOD} detection~\cite{goodfellow2014explaining, hendrycks17baseline, yang2021generalized} has attracted much attention from the research community due to its practical implications, but almost all recent work has primarily considered \ac{OOD} in standard supervised settings with a large number of samples. 
These many examples provide much information, \eg, clear boundaries, for distinguishing \ac{IND} samples from future \ac{OOD} inputs.
However, access to abundant \ac{IND} data is not always available in many settings, necessitating algorithms that can learn from few examples.
Such few-shot settings~\cite{wang2020generalizing, bendre2020learning} commonly arise where data is expensive or difficult to come by~\cite{altae2017low,liang2022few}, quick adaptation to new classes is necessary~\cite{vartak2017meta}, or models are personalized to users~\cite{grauman2022ego4d}.

\begin{figure}[t]
    \centering
    \includegraphics[width=\columnwidth]{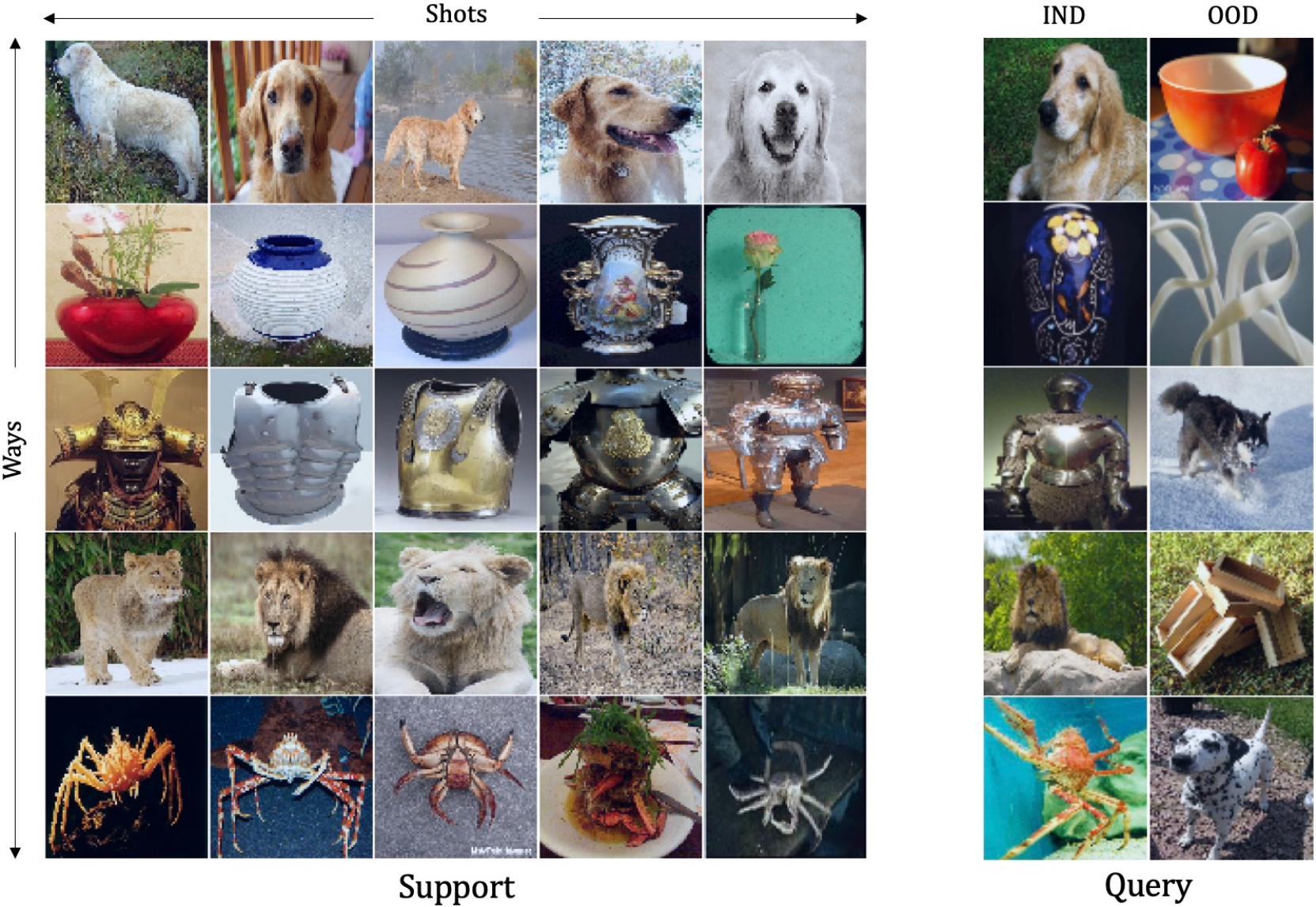}
    \caption{
    \textbf{Few-shot OOD detection and classification.} We show a 5-way 5-shot support set and 10 query examples, from MiniImageNet~\cite{vinyals2016Matching}. The goal is to correctly classify the IND samples (left column) while also detecting the OOD samples (right column).
    }
    \label{fig:fs_ood}
\end{figure}

Although few-shot \ac{OOD} (FS-OOD) detection (Figure~\ref{fig:fs_ood}) is a natural fit for many real-world scenarios~\cite{zheng2016person, grauman2022ego4d, inkawhich2021training}, the intersection of these two problems is relatively underexplored, and solutions are not entirely straightforward as FSL and OOD each exacerbate some of the major challenges of the other:
In few-shot learning, methods are generally starved for generalizability due to the few examples to learn from~\cite{chen2019multi, li2020adversarial}, but in an open-world setting, one must also be wary of over-generalizing, which risks misclassifying \ac{OOD} samples as one of the \ac{IND} classes.
Conversely, for \ac{OOD} detection, having a limited number of samples can handicap learning the \ac{IND} data distribution, making it significantly more challenging to detect those that are \ac{OOD}, especially since neither the \ac{IND} classes nor \ac{OOD} classes are known ahead of time during offline meta-training.
As we empirically show, popular few-shot and OOD methods tend to struggle in the FS-OOD setting.
In particular, we find that many recent state-of-the-art \ac{OOD} methods fail to outperform the simple baseline of maximum softmax predictive probability (MSP)~\cite{hendrycks17baseline} in the few-shot setting.

To address this gap, we develop a novel framework for few-shot \ac{OOD} detection that brings together strategies from few-shot learning and \ac{OOD} detection approaches.
In particular, we adopt a hypernetwork-based~\cite{ha2017hypernetworks, bertinetto2016learning} approach that we call HyperMix. 
HyperMix utilizes Mixup~\cite{zhang2018mixup} to augment the samples in a few-shot episode in two distinct ways.
While Mixup has been used to perform augmentations in the input~\cite{zhang2018mixup} or feature~\cite{verma2019manifold} space, we propose Mixup in the {\em parameter space of hypernetwork-generated classifier weights}.
We show that this greatly improves generalization and OOD detection in few-shot settings.

In addition, previous \ac{OOD} methods found that exposure to \ac{OOD} samples during training can be helpful~\cite{hendrycks2018deep, fort2021exploring, zhang2021fine}. Such methods, however, risk overfitting to the chosen \ac{OOD} set and make the assumption of a relevant \ac{OOD} set being available during training. This assumption is at odds with the unpredictable nature of \ac{OOD} samples in the real world~\cite{lee2018training, yang2021generalized}.
Instead, we show that we can naturally incorporate outlier exposure through the meta-learning process common to few-shot methods~\cite{vinyals2016Matching, snell2017Proto, sung2018learning}, {\em without an additional outlier dataset}: classes not sampled during a particular meta-training task can serve as \ac{OOD} samples.

We rigorously test our proposed methods on standard FSL benchmarks CIFAR-FS~\cite{krizhevsky2009learning, bertinetto2019meta} and MiniImageNet~\cite{deng2009imagenet, vinyals2016Matching}, but in an OOD setting. Our results clearly show that the method we propose outperforms popular \ac{OOD} methods, as well as the MSP baseline, in FS-OOD settings. We will release the code to promote the reproduction of our results.  

In summary, we make the following contributions:
\begin{enumerate}
    \item We empirically show that the current state-of-the-art OOD detection methods have limited success in the few-shot learning scenario: the baseline MSP detector~\cite{hendrycks17baseline} and a related variant~\cite{liang2018enhancing} are more reliable than other OOD methods in few-shot settings.
    \item We propose a support-set augmentation for hypernetworks that mixes generated weight parameters and out-of-episode (OOE) Mixup~\cite{zhang2018mixup} samples to supplement the query set. We name the combination of these two techniques HyperMix.
    \item Through extensive experiments on CIFAR-FS and MiniImageNet, we show that our proposed HyperMix leads to improved FS-OOD detection. 
\end{enumerate}

\section{Related Work}
\vspace{-1mm}
\minisection{\acf{OOD} Detection} As artificial intelligence is increasingly being deployed for real-world use cases, understanding failure modes has become critical for the safety of such systems~\cite{amodei2016concrete, wang2019ai, liang2019toward, badue2021self}.
One important axis is how such systems react when presented with unknown inputs~\cite{yang2021generalized}.
Adversarial examples~\cite{szegedy2013intriguing, goodfellow2014explaining, kurakin2018adversarial} have infamously demonstrated the ability to induce arbitrary responses in neural networks, but even untampered, normal examples can induce unexpected responses when not from the training distribution~\cite{nguyen2015posterior, hein2019relu}.
While machine learning models do tend to be overconfident on \ac{OOD} samples, the confidence on such samples is often lower than that of in-distribution samples and thus can be used as a simple way of identifying \ac{OOD} samples~\cite{hendrycks17baseline}.
Subsequent methods have sought to improve \ac{OOD} detection with more sophisticated strategies, such as learning class-specific mean and covariance of embeddings and then using the Mahalanobis distance~\cite{lee2018simple}.
ODIN~\cite{liang2018enhancing} uses small perturbations and temperature scaling to further separate IND and OOD score distributions.
pNML~\cite{bibas2021single} proposed to use a specific generalization error, the predictive normalized maximum likelihood regret, as the confidence score for OOD detection. 

Outlier exposure has also been a popular technique: by showing the model \ac{OOD} examples and explicitly optimizing for lower confidence, these models are often more robust to outliers.
Some approaches source outliers from other datasets~\cite{dhamija2018reducing, hendrycks2018deep, zhang2021fine}, while others generate them~\cite{lee2018training, vernekar2019out}.
However, the outlier distribution is often unknown in advance, and outliers exposed during training, whether from other datasets or synthesized, may not be representative.

\minisection{Few-Shot Learning (FSL)} 
While many deep learning advancements were enabled by large supervised datasets~\cite{deng2009imagenet, krizhevsky2012imagenet}, for many problems, labeled data can be scarce or expensive; being able to learn from few samples has thus been an active field of research recently.
We direct interested readers to surveys~\cite{wang2020generalizing, bendre2020learning} for a more thorough treatment, highlighting several relevant works here.

Many few-shot methods are metric-based, which seek to learn an embedding such that classification can be done by comparing query and support samples in the learned feature space.
A prominent example of this are Prototypical Networks~\cite{snell2017Proto}, which meta-learns a feature extractor such that the mean of the support sample embeddings for a class can be used for Euclidean distance nearest neighbors. 
Mahalanobis distance~\cite{bateni2020improved}, Earth
Mover’s Distance~\cite{zhang2020deepemd}, and cosine similarity~\cite{chen2019closer} have also been utilized.
Alternatively, methods such as RelationNet~\cite{sung2018learning} learn a metric for comparing samples.
POODLE~\cite{le2021poodle} leverages OOD samples to help the model ignore irrelevant features.

Approaches with hypernetworks~\cite{ha2017hypernetworks}--auxiliary models that generate weights for the primary task model--have also been popular for FSL.
Learnnet~\cite{bertinetto2016learning} proposed generating full weights of a deep discriminitive model from a single sample, scaling parameters by factorizing linear and convolutional layers.
Hypernetworks have subsequently appeared in several FSL works, such as using task embeddings to influence model weights~\cite{wang2019tafe, oreshkin2018tadam}, generating task-specific classifiers on top of a pre-trained encoder~\cite{rangrej2021revisiting, sendera2022hypershot}, or synthesizing convolutional filters for object detection~\cite{perez2020incremental, yin2022sylph}.

\begin{figure*}[t]
    \centering
    \includegraphics[width=0.8\textwidth]{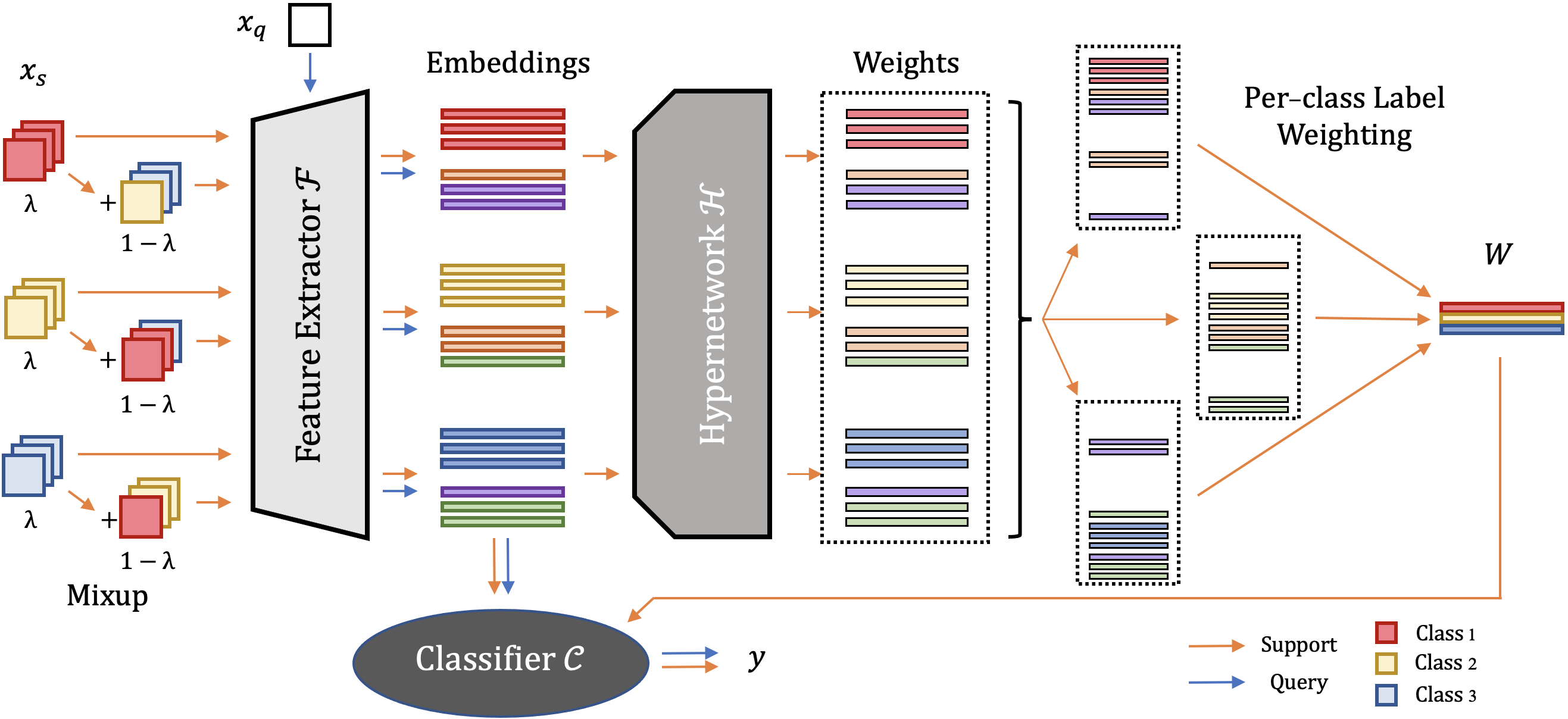}
    \caption{
    A diagram of the proposed ParamMix, for a 3-way 3-shot example episode.
    \textit{Support samples (orange path)}: To generate the classifier, each of the $KN$ inputs $x_s$ (with label $y_s$ denoted by color) in the support set are fed to the feature extractor $\mathcal F$; Mixup-augmented samples are also processed, resulting in $2KN$ embeddings.
    Each embedding is then passed to hypernetwork $\mathcal H$ to generate a corresponding classifier weight.
    The final $N$ classifier weights $W$ are calculated by weighted aggregation based on the samples' Mixup labels (Equation~\ref{eq:parammix_weight}).
    \textit{Query samples (blue path)}: To perform inference, we compose the feature extractor and classifier to produce the prediction $y = \mathcal C(\mathcal F(x_q))$, using $W$ previously generated from the support set.
    }
    \label{fig:parammix}
\end{figure*}

\minisection{Few-Shot \ac{OOD} Detection} 
Few attempts have been made to solve the challenging problem of OOD detection under the few-shot constraint until recently. FS-OOD \cite{wang2020fsood} introduced Out-of-Episode Classifier (OEC) that utilizes examples from the same dataset but not the current episode and showed its effectiveness in few-shot OOD detection. OOD-MAML \cite{Jeong2020oodmaml} incorporated model-agnostic meta-learning (MAML) \cite{finn2017model}, a popular meta-learning approach for FSL, and proposed to synthesize OOD examples from unknown classes via gradient updating of special meta-parameters to help the model learn a sharper decision boundary. FROB \cite{dionelis2022frob} performed self-supervised few-shot negative data augmentation on the distribution confidence boundary and combined it with outlier exposure.

\section{Background}
\label{sec:background}
\vspace{-1mm}
\textbf{OOD Detection} methods seek to identify unfamiliar OOD inputs during inference.
More precisely, \ac{OOD} methods~\cite{hendrycks17baseline} commonly frame the problem as training a model on ``in-distribution'' data $\{x_{IND}, y_{IND}\} \sim \mathcal D_{IND}$, but during test, inputs are drawn from an expanded distribution $\mathcal D_{test} = \mathcal D_{IND} \cup \mathcal D_{OOD}$, with $\mathcal D_{OOD}$ being the out-of-distribution data. 
Generally, the OOD distribution is unknown during training, though some methods utilizing outlier exposure~\cite{hendrycks2018deep} may assume access to an auxiliary dataset of OOD samples $\hat{\mathcal D}_{OOD}$ during training, which may be distinct from the inference time outliers $\mathcal D_{OOD}$.
Samples from this dataset $\hat{x} \sim \hat{\mathcal D}_{OOD}$ are used to augment the IND samples during training.
A common label to assign such OOD samples is one of maximum entropy: $p(\hat{y} | \hat{x}) = 1/N$ for an $N$-way classification problem.

\textbf{Few-Shot Classification} is typically framed as a $K$-shot $N$-way classification problem, where the goal is to leverage $K \times N$ labeled examples from $N$ unseen classes to learn a model $\mathcal{M}$ that can classify $Q$ unlabeled examples from the same classes. We denote the labeled $K \times N$ examples as $\mathcal{D}_s = \{x_s, y_s\}_{s=1}^{KN}$ and the $Q$ unlabeled examples as $\mathcal{D}_q = \{x_q\}_{q=1}^Q$. The labeled $\mathcal{D}_s$ and the unlabeled $\mathcal{D}_q$ examples are often referred to as support and query examples of the \ac{FSC} task. 

To produce an effective model $\mathcal{M}^*$ for novel classes $C_n$, many \ac{FSC} methods~\cite{vinyals2016Matching,snell2017Proto} meta-train $\mathcal{M}$ on many $K$-shot $N$-way classification tasks sampled from a large labeled dataset of base classes $C_b$, which are assumed to be distinct from the classes in $C_n$, \ie $C_n \cap C_b = \emptyset$. Each task sampled from the training dataset of base classes is referred to as an \ac{FSC} episode, which contains a support and query set. Specifically, for episode $i$, $\{\mathcal{D}_s^{(i)}, \mathcal{D}_q^{(i)}\} \sim \mathcal{D}_{train}$, where $\mathcal{D}_{train}$ contains labeled examples from classes $C_b$, and $\{\mathcal{D}_s^{(i)}, \mathcal{D}_q^{(i)}\}$ contain labeled examples from classes $C_b^{(i)} \subset C_b$ and $|C_b^{(i)}| = N$. The optimal classifier is produced by solving the following optimization problem:
\begin{small}
\begin{align}
    \mathcal{M}^* = \argmin_{\mathcal{M}} \Ebb_{\{
    \mathcal{D}_s^{(i)}, \mathcal{D}_q^{(i)}
    \}} &\Big[\mathcal{L}_{EP}\left( \mathcal{M}(\mathcal{D}_s^{(i)}); \mathcal{D}_q^{(i)} \right) \Big], \\
\textrm{where } \mathcal{L}_{EP}\left( \mathcal{M}(\mathcal{D}_s^{(i)}); \mathcal{D}_q^{(i)} \right) &= \Ebb_{D_q^{(i)}} \left[ \mathcal{L}\left(\hat{y}_q^{(i)}, y_q^{(i)}\right)\right].
\end{align}
\end{small}
\noindent Here the prediction $\hat{y}_q^{(i)} = \mathcal{M}(\mathcal{D}_s^{(i)})(x_q^{(i)})$ and $\mathcal{L}_{EP}$ corresponds to the loss for the $i^{th}$ episode.

\textbf{Mixup}~\cite{zhang2018mixup} is a data augmentation technique aiming to linearize model behaviour between samples. Given a pair of inputs $(x_1, y_1), (x_2, y_2) \sim \mathcal{D}$ drawn from the data distribution, Mixup constructs an augmented sample $(\tilde{x}_1, \tilde{y}_1)$ as a convex combinations of the two inputs and labels:
\begin{align}
    \tilde{x} &=  \lambda ~ x_1 + (1-\lambda) ~ x_2 \label{eq:mixup_x} \\
    \tilde{y} &=  \lambda ~ y_1 + (1-\lambda) ~ y_2 \label{eq:mixup_y}
\end{align}
with $\lambda \in [0,1]$ being a weighting factor drawn randomly; Uniform and Beta distributions are common.
These augmented samples are then used as additional training data.

\section{Methods}
\vspace{-1mm}
\subsection{FS-OOD Classification and Detection}
\vspace{-1mm}
We formulate FS-OOD detection and classification as the intersection of the OOD detection and few-shot classification problems.
Given an $N$-way $K$-shot support set $\mathcal{D}_s = \{x_s, y_s\}_{s=1}^{KN}$ (in-distribution data $\mathcal D_{IND}$) and query samples $\mathcal{D}_q = \{x_q\}_{q=1}^Q$, where $x_q$ is drawn from either the $N$ classes or the OOD distribution $\mathcal D_{OOD}$, the goal is to classify each query $x_q$ as one of the $N$ classes or as OOD.

As FSC aims to generalize to new tasks during meta-test, the FS-OOD detection problem poses a notable challenge over standard OOD detection in that the IND data distribution $\mathcal D_{IND}$ is unknown during meta-training, in addition to the usually unknown OOD distribution $\mathcal D_{OOD}$.
Instead, a model that can quickly adapt to a new IND distribution given a few samples must be learned.

\vspace{-1mm}
\subsection{Hypernetwork Framework}
\label{sec:hypernetwork}
\vspace{-1mm}
Because of their strong adaptation capabilities~\cite{galanti2020modularity, rangrej2021revisiting}, we adopt a hypernetwork~\cite{ha2017hypernetworks} framework for FS-OOD: We instantiate model $\mathcal{M}$ as a feature extractor $\mathcal{F}$ and a linear classifier $\mathcal{C}$ parameterized by $W$. Following previous works~\cite{chen2019closer}, we first pre-train $\mathcal{F}$ on the dataset of base classes $\mathcal D_b$. Rather than directly learning $W$ by gradient descent, we instead learn a hypernetwork $\mathcal{H}$ that generates the parameters for $\mathcal{C}$ from the support set. 
A hypernetwork-based model allows us to leverage the \textit{learning to learn} paradigm, training a universal architecture model which can be quickly adapted to a new task by generating task-specific weights. Unlike fine-tuning and second-order procedures (inner and outer loop), hypernetworks are ``training-free'' when producing a classifier for a new task, resulting in faster adaptation and lower computational requirements. 

\minisection{Meta-training of $\mathcal{H}$} The hypernetwork $\mathcal{H}$ is trained on many classification tasks sampled from the dataset of base classes $C_b$ in the meta-training stage. This meta-training is done in an episodic fashion by drawing $M$ few-shot tasks from $\mathcal{D}_{train}$. For each episode $i$, the hypernetwork takes the support embeddings extracted using $\mathcal F$ and generates sample-specific weight codes $\mathcal{H}(\mathcal{F}(x_s^{(i)}))$, which are used to generate the classifier weights as follows:
\begin{align}
    w_n = \frac{1}{K} \sum_{s=1}^{KN}\onev_{[y_s=n]}\mathcal{H}\left(\mathcal{F}(x_s^{(i)})\right). \label{eq:weight_gen}
\end{align}
Let the classifier weights derived from the support set of episode $i$ be $W(\mathcal{D}_s^{(i)}) = [w_1, \dots, w_n]$. The class-logits for queries are predicted as $\hat{y}_q = \mathcal{C}\left(\mathcal{F} \left(x_q\right)|W(\mathcal{D}_s^{(i)})\right)$. The loss function at the meta-training stage is as follows:
\begin{small}
\begin{align}
    \mathcal{H}^* = \argmin_{\mathcal{H}}\, \Ebb_{\{\mathcal{D}_s^{(i)}, \mathcal{D}_q^{(i)}\}} &\left[ \mathcal{L}_{EP} \left(W(\mathcal{D}_s^{(i)}); \mathcal{D}_q^{(i)} \right)\right], \label{eq:hypernetwork_training} \\
    \textrm{where } \mathcal{L}_{EP} \left(W(\mathcal{D}_s^{(i)}); \mathcal{D}_q^{(i)}\right) &= \Ebb_{\mathcal{D}_q^{(i)}}\left[\mathcal{L}_{CCE}\left(\hat{y}_q^{(i)}, y_q^{(i)}\right)\right] \nonumber \\
    \textrm{and } \mathcal{L}_{CCE}\left(\hat{y}_q^{(i)}, y_q^{(i)}\right) = &-y_q^{(i)} \cdot \log{( \hat{y}_q^{(i)})} \nonumber
\end{align}
\end{small}
\noindent is the Categorical Cross Entropy (CCE) loss. Note, the loss gradient in (\ref{eq:hypernetwork_training}) can also be backpropagated into $\mathcal F$ to fine-tune the feature extractor, providing some improvement.

\minisection{OOD Detection during Inference} After meta-training the model, we use the maximum softmax probability~\cite{hendrycks17baseline} as the score for the OOD detection task during inference. In particular, we define the score $s(x_q)$ for a given query $x_q$ at meta-testing as $s(x_q) = \max_{c}{\left[\,\hat{y}_{q,c}\,\right]}$, where  $\hat{y}_{q,c}$ is the $c^{th}$ component of the softmax vector $\hat{y}_q = \mathcal{C}\left(\mathcal{F} \left(x_q\right)|W\right)$. If $s(x_q) < \tau$, with $\tau$ being the OOD threshold, we call the query an \ac{OOD} sample. However, note that the OOD detection task is evaluated using metrics that alleviate the need to set a fixed threshold $\tau$ (see Section~\ref{sec:setup}).

\vspace{-1mm}
\subsection{HyperMix}
\label{sec:hypermix}
\vspace{-2mm}
We find our hypernetwork framework for FS-OOD outperforms other common few-shot approaches, but such a set-up alone does not incorporate any OOD detection capabilities.
We thus propose HyperMix: a support and query set augmentation technique based on Mixup~\cite{zhang2018mixup} for training the hypernetwork during the meta-training stage. 
In contrast to the existing Mixup data augmentation technique, which uses linear Mixup of the model inputs and target labels, HyperMix uses Mixup in two novel ways uniquely adapted to our hypernetwork framework and FS-OOD settings, which we call ParamMix and OOE-Mix.  

\minisection{ParamMix}  
ParamMix augments the support set using Mixup and uses weighted aggregation to compute the classifier weights. 
Let the mixed support samples and the corresponding label be denoted as $\tilde{x}_s$ and $\tilde{y}_s$ respectively. 
The mixing parameter $\lambda_{PM}$ is sampled from a fixed Beta distribution, and the extracted parameters of the classifiers are calculated using weighted aggregation of sample-specific weight codes, where the aggregation weights are determined based on the probabilities of mixed samples. ParamMix modifies the weight generation in (\ref{eq:weight_gen}) for soft labels:
\begin{align}
    w_n = \frac{1}{\sum_{s=1}^{S} \tilde{y}^{(s)}_n} \sum_{s=1}^{S} \tilde{y}^{(s)}_n \enspace \mathcal{H}\left(\mathcal{F}\left(\tilde{x}_s\right)\right). \label{eq:parammix_weight}
\end{align}
where $\tilde{x}_s = \lambda_{PM} \enspace x_s^{(1)} + (1 - \lambda_{PM}) \enspace x_s^{(2)}$ is a mixed sample formed from two support set images, $\tilde{y}^{(s)}_n$ corresponds to the probability of $\tilde{x}_s$ belonging to the label $n$, and $S$ is the total number of support samples after augmentation with Mixup.
We summarize ParamMix in Figure~\ref{fig:parammix}.

\minisection{OOE-Mix} 
Outlier exposure~\cite{hendrycks2018deep, fort2021exploring, zhang2021fine} has been shown to be highly effective for OOD detection, but it often assumes access to a dataset of outliers, which is often practical.
On the other hand, we note that the meta-learning process of sampling a subset of the base classes $C_b^{(i)}$ to construct a few-shot episode means there are classes $C_b \setminus C_b^{(i)}$ available that are not part of the $N$-way classification for episode $i$.
We refer to the samples from $C_b^{(i)}$ as in-episode (INE) and samples from $C_b \setminus C_b^{(i)}$ as out-of-episode (OOE).
As the OOE samples are from the same dataset, not only are these samples representative and of the same domain as the INE data, but the mutually exclusive nature of the labels means that they are also not in-distribution.
We can thus consider the ample samples from $C_b \setminus C_b^{(i)}$ as realistic ``near'' outliers~\cite{winkens2020contrastive} for this particular few-shot episode. 

\begin{figure}[t]
    \centering
    \includegraphics[width=\columnwidth]{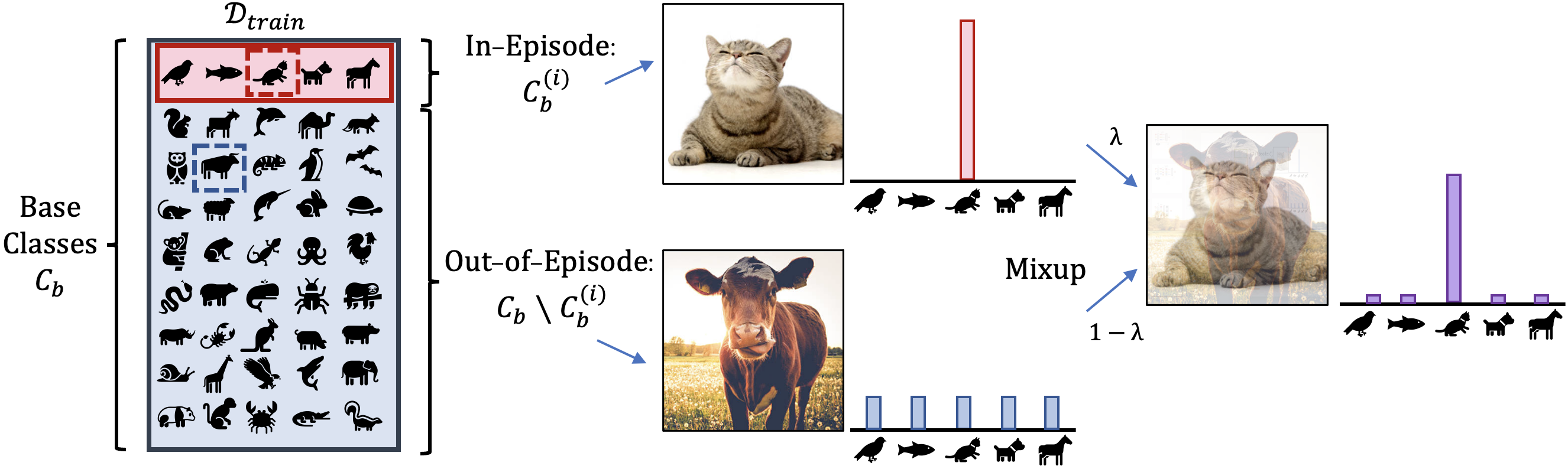}
    \caption{
    Illustration of OOE-Mix augmentation process. 
    For every few-shot episode during meta-training, $N$ classes $C_b^{(i)}$ are sampled from the base classes $C_b$.
    The remaining out-of-episode (OOE) classes can be used as a source for realistic near \ac{OOD} examples.
    We mix these OOE samples with the sampled in-episode (INE) examples to improve generalization.
    }
    \label{fig:ooemix}
    \vspace{-3mm}
\end{figure}

While we can expose OOE samples during meta-training as is, we find mixing INE and OOE samples to be highly effective. 
Therefore, we propose OOE-Mix to augment the query set with mixed INE and OOE samples (Figure~\ref{fig:ooemix}). Specifically, we sample INE and OOE samples as:
\begin{align}
    x^{(i)}_1 \sim \mathcal{D}^{(i)}_{INE};& \quad x^{(i)}_2 \sim \mathcal{D}^{(i)}_{OOE}; \quad p(y | x^{(i)}_2) = 1 / N
\end{align}
As we consider the OOE samples to be OOD, we assign OOE samples with a uniform distribution label, to enforce maximal uncertainty.
Similar to Equations~\ref{eq:mixup_x} and \ref{eq:mixup_y}, we then apply Mixup, but specifically between pairs of INE and OOE samples:
\begin{align}
    \tilde{x}^{(i)} &=  \lambda_{OM} ~ x^{(i)}_1 + (1-\lambda_{OM}) ~ x^{(i)}_2 \\
    p(y | \tilde{x}^{(i)}) &=  \lambda_{OM}~ p(y | x^{(i)}_1) + (1-\lambda_{OM})~p(y | x^{(i)}_2)
\end{align}
with $\lambda_{OM} \sim \text{Beta}(a_{OM}, b_{OM})$. We augment the query dataset with the generated mixed samples while optimizing the loss function in Equation~(\ref{eq:hypernetwork_training}).
We find this helpful as the addition of Mixup between INE and OOE samples encourages smooth behaviour in the classifier's decision space.
This helps with generalization, a critical component for few-shot settings.

\vspace{-1mm}
\section{Experiments}
\vspace{-1mm}
\subsection{Datasets}
\vspace{-1mm}
We run experiments on popular few-shot learning datasets CIFAR-FS~\cite{bertinetto2019meta} and MiniImageNet~\cite{vinyals2016Matching}, but we adapt them for few-shot OOD detection. 
In particular, for each meta-testing task, we include query images from the unsampled meta-test classes (those not in the $N$-way task) as OOD inputs (see Evaluation Protocol in Section~\ref{sec:setup}).

\textbf{CIFAR-FS} consists of $32 \times 32$ color images from the CIFAR-100 dataset~\cite{krizhevsky2009learning}, split into 64, 16, and 20 classes for meta-training, meta-validation, and meta-testing, with 600 images per class.

\textbf{MiniImageNet}~\cite{vinyals2016Matching} is a common few-shot benchmark consisting of $84 \times 84$ color images, with 100 classes of 600 examples each. Similar to CIFAR-FS, we use 64 classes for meta-training, 16 for validation, and 20 for meta-testing.

\vspace{-1mm}
\subsection{Set-up}
\label{sec:setup}
\vspace{-1mm}
Our experimental settings closely resemble the traditional few-shot classification benchmarks, with the primary difference being that half the meta-test query samples are drawn from outside of the $N$-way classification problem, \ie OOD samples. We use 5-shot and 10-shot settings during the meta-training phase. All experiments were done on a GeForce GTX 1080 GPU with 12GB memory.

\begin{table*}[t]
  \centering
  \caption{\textbf{FS-OOD Classification and Detection.} *OE is the outlier exposure method and thus requires auxiliary data, which we do not assume in our experiments; we instead use OOE sampling as the auxiliary outliers. Best value shown in \textbf{bold}; second best shown in \textit{italics}.}
  \resizebox{0.85\textwidth}{!}{
    \begin{tabular}{cl||ccc|ccc}
    \toprule
    && \multicolumn{3}{|c|}{5-shot 5-way} & \multicolumn{3}{c}{10-shot 5-way}\\
    \midrule
    & \multicolumn{1}{l||}{Method} &
    \multicolumn{1}{c}{IND Acc $\uparrow$} &
    \multicolumn{1}{c}{AUROC $\uparrow$} & \multicolumn{1}{c|}{FPR@90 $\downarrow$} &
    \multicolumn{1}{c}{IND Acc $\uparrow$} & \multicolumn{1}{c}{AUROC $\uparrow$} & \multicolumn{1}{c}{FPR@90 $\downarrow$}\\
    \midrule
    && \multicolumn{6}{c}{FS-CIFAR-100~\cite{bertinetto2019meta}} \\ 
    \midrule
    & Fine-tune w/ MSP~\cite{hendrycks17baseline} & 80.38 $\pm$ 0.41 & 72.00 $\pm$ 0.40 & 75.20 $\pm$ 0.68 & \textbf{85.04 $\pm$ 0.72} & 74.30 $\pm$ 0.81 & 71.25 $\pm$ 1.43 \\
    \midrule
    {\multirow{4}{*}{\rotatebox[origin=c]{90}{ProtoNet}}}
    & MSP~\cite{hendrycks17baseline} & 80.09 $\pm$ 0.42 & 75.42 $\pm$ 0.37 & 66.42 $\pm$ 0.72 & 83.56 $\pm$ 0.76 & 75.55 $\pm$ 0.75 & 67.72 $\pm$ 1.46 \\
    & ODIN~\cite{liang2018enhancing} & 80.09 $\pm$ 0.42& 75.31 $\pm$ 0.37 & 66.69 $\pm$ 0.72 & 83.56 $\pm$ 0.76 & 75.39 $\pm$ 0.75  & 67.90 $\pm$ 1.44 \\
    & DM~\cite{lee2018simple} & 80.09 $\pm$ 0.42& 66.24 $\pm$ 1.14 & 77.43 $\pm$ 1.47 & 83.56 $\pm$ 0.76 & 68.03 $\pm$ 0.71 & 79.69 $\pm$ 1.20\\
    & OEC~\cite{wang2020few} & 76.45 $\pm$ 0.93 & 74.32 $\pm$ 0.94 &  67.84 $\pm$ 0.81 & 80.86 $\pm$ 0.78 & 74.40 $\pm$ 0.77 & 66.00 $\pm$ 1.48\\
    & OE*~\cite{hendrycks2018deep} & 78.31 $\pm$ 0.41 & 73.13 $\pm$ 0.45 & 68.61 $\pm$ 0.94 & 81.01 $\pm$ 0.82 & 74.14 $\pm$ 0.91 & 68.68 $\pm$ 1.01 \\
    \midrule
    {\multirow{10}{*}{\rotatebox[origin=c]{90}{Hypernetwork}}}
    & MSP~\cite{hendrycks17baseline} & 78.89 $\pm$ 0.45 & 78.38 $\pm$ 0.39 & 57.65 $\pm$ 0.78 & 81.58 $\pm$ 0.41 & 81.42 $\pm$ 0.35 & 54.19 $\pm$ 0.79 \\
    & Entropy~\cite{mukhoti2021deterministic} & 78.89 $\pm$ 0.45 & 69.67 $\pm$ 0.43 & 70.33 $\pm$ 0.70 & 81.58 $\pm$ 0.41 & 71.92 $\pm$ 0.41 & 69.02 $\pm$ 0.71\\
    & ODIN~\cite{liang2018enhancing} & 78.89 $\pm$ 0.45 & 78.73 $\pm$ 0.38 & 56.42 $\pm$ 0.78 & 81.58 $\pm$ 0.41 & 81.83 $\pm$ 0.35 & 52.95 $\pm$ 0.79\\
    & DM~\cite{lee2018simple} & 78.89 $\pm$ 0.45 &63.67 $\pm$ 0.22 &80.06 $\pm$ 0.29 & 81.58 $\pm$ 0.41 & 67.48 $\pm$ 0.18 & 76.23 $\pm$ 0.27\\
    & pNML~\cite{bibas2021single} & 78.89 $\pm$ 0.45 & 60.41 $\pm$ 0.49 & 78.53 $\pm$ 0.65 & 81.58 $\pm$ 0.41 & 66.43 $\pm$ 0.42 & 72.56 $\pm$ 0.46\\
    & OEC~\cite{wang2020few} & 77.51 $\pm$ 0.44 & 79.11 $\pm$ 0.34 &  56.86 $\pm$ 0.79 & 80.61 $\pm$ 0.61 & 82.06 $\pm$ 0.41 & 53.19 $\pm$ 0.74\\
    & OE*~\cite{hendrycks2018deep} & 79.46 $\pm$ 0.42 & \textit{81.68 $\pm$ 0.35} & \textit{55.21 $\pm$ 0.80} & 81.69 $\pm$ 0.54 & 83.14 $\pm$ 0.33 & 52.16 $\pm$ 0.79\\
    \arrayrulecolor{gray}\cmidrule{2-8}\arrayrulecolor{black}
    & ParamMix (ours) & 78.45 $\pm$ 0.49 & 79.24 $\pm$ 0.37 & 57.12 $\pm$ 0.77 & 81.22 $\pm$ 0.42 & 82.23 $\pm$ 0.35 & 52.33 $\pm$ 0.78 \\
    & OOE-Mix (ours) &  \textit{80.88 $\pm$ 0.42} & 81.11 $\pm$ 0.35 & 56.38 $\pm$ 0.80 & 83.46 $\pm$ 0.40 & \textbf{84.08 $\pm$ 0.32} & \textit{50.97 $\pm$ 0.81} \\
    & HyperMix (ours) & \textbf{81.33 $\pm$ 0.41} & \textbf{82.43 $\pm$ 0.33} & \textbf{53.69 $\pm$ 0.79} & \textit{83.79 $\pm$ 0.38} & \textit{83.20 $\pm$ 0.34} & \textbf{50.66 $\pm$ 0.80} \\
    \midrule
    && \multicolumn{6}{c}{MiniImageNet~\cite{vinyals2016Matching}} \\ \midrule
    {\multirow{10}{*}{\rotatebox[origin=c]{90}{Hypernetwork}}}
    & MSP~\cite{hendrycks17baseline} & 72.13 $\pm$ 0.47 & 72.70 $\pm$ 0.41 & 69.97 $\pm$ 0.73 & 75.15 $\pm$ 0.44 & 75.24 $\pm$ 0.39 & 66.40 $\pm$ 0.75 \\
    & Entropy~\cite{mukhoti2021deterministic} & 72.13 $\pm$ 0.47 & 62.73 $\pm$ 0.43 & 79.64 $\pm$ 0.59 & 75.15 $\pm$ 0.44 & 71.92 $\pm$ 0.41 & 69.02 $\pm$ 0.71\\
    & ODIN~\cite{liang2018enhancing} & 72.13 $\pm$ 0.47 & 72.86 $\pm$ 0.41 & 68.84 $\pm$ 0.73 & 75.15 $\pm$ 0.44 & 75.45 $\pm$ 0.38 & 65.94 $\pm$ 0.75\\
    & DM~\cite{lee2018simple} & 72.13 $\pm$ 0.47 & 61.00 $\pm$ 0.18 & 83.71 $\pm$ 0.22 & 75.15 $\pm$ 0.44 & 63.58 $\pm$ 0.18 & 81.71 $\pm$ 0.24\\
    & pNML~\cite{bibas2021single} & 72.13 $\pm$ 0.47 & 56.32 $\pm$ 0.34 & 85.94 $\pm$ 0.46 & 75.15 $\pm$ 0.44 & 62.11 $\pm$ 0.42 & 83.15 $\pm$ 0.59\\
    & OEC~\cite{wang2020few} & 71.11 $\pm$ 0.51 & 72.81 $\pm$ 0.36 & 68.61 $\pm$ 0.71 & 73.94 $\pm$ 0.40 & 75.01 $\pm$ 0.44 & 64.94 $\pm$ 0.71\\
    & OE*~\cite{hendrycks2018deep} & 72.64 $\pm$ 0.44 & 73.81 $\pm$ 0.39 & 68.78 $\pm$ 0.73 & 75.81 $\pm$ 0.43 & 74.84 $\pm$ 0.40 & 65.46 $\pm$ 0.74\\
    \arrayrulecolor{gray}\cmidrule{2-8}\arrayrulecolor{black}
    & ParamMix (ours) & 71.52 $\pm$ 0.45 & 73.87 $\pm$ 0.40 & 69.14 $\pm$ 0.73 & 74.99 $\pm$ 0.44 & 75.46 $\pm$ 0.39 & 65.69 $\pm$ 0.75 \\
    & OOE-Mix (ours) & \textit{73.95 $\pm$ 0.45} & \textit{74.75 $\pm$ 0.39 }& \textit{66.62 $\pm$ 0.74} & \textbf{78.25 $\pm$ 0.41} & \textit{76.66 $\pm$ 0.38} & \textit{64.55 $\pm$ 0.77} \\
    & HyperMix (ours) & \textbf{74.38 $\pm$ 0.41} & \textbf{75.03 $\pm$ 0.43} & \textbf{65.57 $\pm$ 0.75} & \textit{77.81 $\pm$ 0.42} & \textbf{77.50 $\pm$ 0.38} & \textbf{63.15 $\pm$ 0.76}
    \\
    \bottomrule
    \end{tabular}
}
\label{tab:fs_ood}%
\end{table*}%

\minisection{Implementation of $\mathcal{F}$} We implement the feature extractor $\mathcal{F}$ as a ResNet-12~\cite{he2016deep}, which has 4 residual blocks with 3 convolutional layers of 64, 160, 320, and 640 3 × 3 kernels. The first three blocks are followed by $2 \times 2$ average pooling, resulting in features of size 640. 
Inspired by recent self-supervised learning methods for few-shot classification~\cite{Gidaris2019BoostingFV, Mangla2020ChartingTR, Rajasegaran2020SelfsupervisedKD}, we train $\mathcal{F}$ using a combination of supervised and self-supervised criteria. In particular, we follow the strategy of \cite{Rajasegaran2020SelfsupervisedKD}, where the image-label pair $\{x, y\}$ is augmented to produce $\{x(r), y, r\}$, where $x(r)$ is the image $x$ rotated by $r$ degrees and $r \in \{0, 90, 180, 270\}$. In the pre-training stage, a stack of two linear classifiers are learned on top of $\mathcal{F}$, where the first classifier predicts the class logits $\hat{y}$ for the $|C_b|$-way classification task, and the second classifier predicts the rotation logits $\hat{r}$. The model is trained using $\mathcal{L}_{CCE}\left(\mathrm{softmax}(\hat{y}), y\right) + \mathcal{L}_{BCE}\left(\mathrm{sigmoid}(\hat{r}), r)\right)$, where $CCE$ and $BCE$ correspond to the Categorical and Binary Cross-Entropy loss. After convergence, the two linear classifiers are discarded~\cite{bordes2022guillotine}, and $\mathcal{F}$ is used to extract image embeddings in the meta-training stage.
During pre-training of $\mathcal{F}$, the batch size was set to 64. Each image in the batch was rotated three times at 90, 180, and 270 degrees, resulting in an effective batch size of 256. We train $\mathcal{F}$ for 200 epochs using Stochastic Gradient Descent (SGD) optimizer with Nesterov momentum~\cite{sutskever2013Nesterov} of 0.9 and weight decay of 5e-4. We start with an initial learning rate of 0.1 and decay by 0.2 at 60, 120, and 160 epochs. 

\minisection{Implementation of $\mathcal{H}$} We parameterize the hypernetwork as a multi-layered perceptron, containing two fully-connected layers of size 256. We generate the classifier weights using the procedure described in Equations~\ref{eq:weight_gen} and \ref{eq:parammix_weight}. Note that while the number of parameters generated in Equation~\ref{eq:weight_gen} corresponds to the number of classes $N$, in practice, we generate $N+1$ parameters and split them into classifier weights and biases. We meta-train $\mathcal{H}$ for 50 epochs with SGD optimizer having a fixed learning rate of 0.001. Each epoch had 200 batches, where each batch contained upto 4 episodes/tasks. The input images are augmented using random crop, color jitter, and horizontal flip.

\minisection{Evaluation Protocol} We sample 400 episodes randomly from the meta-test set containing novel classes $C_n$ that the model has not previously encountered. In each episode, the model is given a labeled support set of $KN$ samples that is referred to as the IND samples, and a set of 20 unlabeled queries. Of the 20 queries, 10 belong to the $N$ IND classes and the remaining 10 are OOD. In our experiments, we consider the \textit{near}-OOD detection task, which is considered a harder problem compared to \textit{far}-OOD detection~\cite{winkens2020contrastive}. Unlike \textit{far}-OOD detection, which uses OOD samples from a different dataset for evaluation, in \textit{near}-OOD detection, we reserve classes from the same dataset as OOD during evaluation. In our experiments, we consider the classes outside a given episode as OOD during evaluation. Note that both IND and OOD samples belong to classes that are not known a priori to the model. We evaluate the model on the few-shot $N$-way classification task for IND queries, and the binary OOD detection task. We use classification accuracy to measure the few-shot classification performance, and AUROC and FPR@90 to measure the OOD detection performance.

\vspace{-1mm}
\subsection{Baselines}
\label{sec:baselines}
\vspace{-2mm}
Few-shot OOD detection and classification is an underexplored problem, meaning there are relatively few previous works.
Standard few-shot methods lack a mechanism to reject outlier samples~\cite{liang2022few}, which leads to misclassifications for all OOD samples.
Thus, we instead primarily compare against several recently proposed OOD methods, with restricted numbers (shots) of IND samples:
\begin{tight_itemize} 
\item Maximum Softmax predictive Probability (MSP)~\cite{hendrycks17baseline}:  MSP uses maximum probability from the model as the \ac{IND} confidence score.
\item Entropy~\cite{mukhoti2021deterministic}:  It uses the predictive entropy of the output probabilities as the \ac{OOD} score.
\item ODIN~\cite{liang2018enhancing}: ODIN also uses MSP, however, it employs temperature scaling and input perturbation based on the gradient of the loss function to increase the margin between maximum probability obtained for IND and OOD samples.
\item Deep Mahalanobis (DM)~\cite{lee2018simple}: DM calculates the class-conditional mean and covariance for each class to calculate the confidence score of query samples at each layer of the encoder. Scores are computed by a logistic regression model learned with validation data.
\item Predictive Normalized Maximum Likelihood (pNML)~\cite{bibas2021single}: pNML uses a generalization error based on predictive normalized maximum likelihood regret as the OOD score.

\item Outlier Exposure (OE)~\cite{hendrycks2018deep}: OE approaches expose OOD samples to the model, trained with a uniform distribution label in the meta-training stage. Note that this assumes the availability of an auxiliary outlier dataset, which we do not use in our experiments. 
Instead, we show OE with our proposed strategy of leveraging OOE samples from $C_b \setminus C_b^{(i)}$.

\item Out-of-Episode Classifier (OEC)~\cite{wang2020few}: OEC is a recently proposed method for FS-OOD detection, which uses binary cross-entropy objective to learn an IND/OOD binary classifier. Similar to OE, we provide OOE samples to OEC during meta-training.

\end{tight_itemize}
We primarily perform our comparisons using the hypernetwork implementation described in Section~\ref{sec:setup}, but we also perform experiments with some of the baselines using fine-tuning~\cite{chen2019closer} and ProtoNet~\cite{snell2017Proto} as the few-shot learning algorithm, to demonstrate the efficacy of our hypernetwork.
We further describe how we adapted the above methods to the FS-OOD setting in Appendix~\ref{apx:baseline_adaptations}. In addition to our proposed HyperMix, we also perform ablations, analyzing the two components of our approach individually:
\begin{tight_itemize} 
\item ParamMix: We apply Mixup to the parameters $W$ generated by hypernetwork $\mathcal{H}$ as described in Section~\ref{sec:hypermix}, but without OOE samples during meta-training.
 
\item OOE-Mix: Similar to OE~\cite{hendrycks2018deep}, we provide OOE samples to the model from the unsampled base classes, and additionally apply Mixup between pairs of INE and OOE samples as described in Section~\ref{sec:hypermix}.
\end{tight_itemize}
Note that our proposed HyperMix is the combination of ParamMix and OOE-Mix. The hyperparameters for all the methods are chosen based on grid search done on validation dataset. The chosen hyperparameters and candidates are discussed in Appendix~\ref{apx:hyperparameters}.

\begin{table}[t]
  \centering
  \caption{ParamMix, OOE-Mix, and Hyper-Mix on mixed samples on CIFAR-FS. Best values shown in \textbf{bold}; second best in \textit{italics}.}
  \vspace{-2mm}
  \resizebox{0.85\columnwidth}{!}{
    \begin{tabular}{l||ccc}
    \toprule
    \multicolumn{1}{l||}{Method} &
    \multicolumn{1}{c}{IND Acc $\uparrow$} &
    \multicolumn{1}{c}{AUROC $\uparrow$} & \multicolumn{1}{c|}{FPR@90 $\downarrow$} \\
    \midrule
    & \multicolumn{3}{c}{Test OOD Dataset: INE-OOE Mixture} \\
    \midrule
    ParamMix & 80.44 $\pm$ 0.86 & 80.61 $\pm$ 0.73 & 55.94 $\pm$ 1.58\\
    OOE-Mix  & \textit{84.24} $\pm$ 0.47  & \textbf{83.47} $\pm$ 0.65 & \textit{52.60} $\pm$ 1.61 \\
    HyperMix & \textbf{84.51} $\pm$ 0.78 & \textit{83.43} $\pm$ 0.66 & \textbf{50.59} $\pm$ 1.60 \\
    \midrule
    & \multicolumn{3}{c}{Test OOD Dataset: INE-INE Mixture} \\
    \midrule
    ParamMix & 81.61 $\pm$ 0.86 & \textbf{65.93} $\pm$ 0.65 & \textbf{83.03} $\pm$ 1.04 \\
    OOE-Mix  &  \textit{84.13} $\pm$ 0.76 & 61.03 $\pm$ 0.61 & 87.86 $\pm$ 0.84 \\
    HyperMix & \textbf{84.46} $\pm$ 0.75 & \textit{62.00} $\pm$ 0.62 & \textit{86.41} $\pm$ 0.89  \\
    \bottomrule
    \end{tabular}
}
\label{tab:mixup_ablation}%
\vspace{-4mm}
\end{table}%

\vspace{-2mm}
\subsection{Few-shot In-Distribution Classification and OOD Detection}
\label{sec:results}
\vspace{-2mm}
We report FS-OOD performance on CIFAR-FS~\cite{bertinetto2019meta} and MiniImageNet~\cite{vinyals2016Matching} in Table~\ref{tab:fs_ood}.
We quantify OOD detection performance with area under the receiver operating characteristic curve (AUROC) and false positive rate at $90\%$ recall rate (FPR@90); we simultaneously also measure classification accuracy of in-distribution samples (IND Acc). 
Note that many of the baselines are post-hoc methods on the same meta-trained model and thus have the same IND accuracy.

\minisection{Choice of Few-shot Framework}
Our proposed hypernetwork framework for FS-OOD outperforms more basic approaches like fine-tuning or ProtoNet. 
For example, for the MSP baseline, a hypernetwork approach improves AUROC and FPR@90 by almost $3\%$ and $9\%$, respectively, compared to ProtoNet on 5-shot 5-way classification on FS-CIFAR-100; this grows to a $6.38\%$ and $17.55\%$ improvement in FS-CIFAR-100 AUROC and FPR@90 compared to MSP with fine-tuning.
Similar improvements to OOD detection can be seen in the 10-shot 5-way case.

\begin{table*}[t]
  \centering
  \caption{\textbf{Noisy FS-OOD Detection (10-shot 5-way).} HyperMix outperforms the baseline methods even in noisy support set settings.}
  \vspace{-2mm}
  \resizebox{0.87\textwidth}{!}{
    \begin{tabular}{l||cc|cc|cc|cc}
    \toprule
    &
    \multicolumn{2}{c|}{10\% Noise} 
    & \multicolumn{2}{c|}{20\% Noise}
    & \multicolumn{2}{c|}{30\% Noise}
    & \multicolumn{2}{c}{40\% Noise}\\
    \midrule
    \multicolumn{1}{l||}{Method} &
    \multicolumn{1}{c}{AUROC $\uparrow$} & \multicolumn{1}{c|}{FPR@90 $\downarrow$} 
    & \multicolumn{1}{c}{AUROC $\uparrow$} & \multicolumn{1}{c|}{FPR@90 $\downarrow$} 
    & \multicolumn{1}{c}{AUROC $\uparrow$} & \multicolumn{1}{c|}{FPR@90 $\downarrow$}
    & \multicolumn{1}{c}{AUROC $\uparrow$} & \multicolumn{1}{c}{FPR@90 $\downarrow$}
    \\
    \midrule
    & \multicolumn{8}{c}{FS-CIFAR-100~\cite{bertinetto2019meta}} \\ \midrule
    MSP~\cite{hendrycks17baseline} & 79.93 $\pm$ 0.36 & 57.15 $\pm$ 0.79 & 77.35 $\pm$ 0.38 & 62.28 $\pm$ 0.78 & 72.17 $\pm$ 0.42 & 70.11 $\pm$ 0.73 & 63.88 $\pm$ 0.47 & 78.42 $\pm$ 0.62\\
    Entropy~\cite{mukhoti2021deterministic} & 71.61 $\pm$ 0.83 & 69.16 $\pm$ 1.40 & 70.68 $\pm$ 0.41 & 71.92 $\pm$ 0.70 & 67.93 $\pm$ 0.43 & 76.56 $\pm$ 0.66 & 63.28 $\pm$ 0.45 & 81.49 $\pm$ 0.59\\
    ODIN~\cite{liang2018enhancing} & 80.33 $\pm$ 0.36 & 55.69 $\pm$ 0.78 & 77.79 $\pm$ 0.38 & 60.88 $\pm$ 0.77 & 72.56 $\pm$ 0.42 & 68.98 $\pm$ 0.73 & 64.08 $\pm$ 0.46 & 77.73 $\pm$ 0.63\\
    DM~\cite{lee2018simple} & 63.52 $\pm$ 0.18 & 80.92 $\pm$ 0.24 & 60.33 $\pm$ 0.18 & 83.63 $\pm$ 0.22 & 57.40 $\pm$ 0.18 & 85.60 $\pm$ 0.20 & 54.79 $\pm$ 0.18 & 87.11 $\pm$ 0.19 \\
    pNML~\cite{bibas2021single} & 62.77 $\pm$ 0.91 & 74.40 $\pm$ 1.09 & 60.64 $\pm$ 0.93 & 75.56 $\pm$ 1.12 & 59.81 $\pm$ 0.92 &  78.61 $\pm$ 1.04 & 57.12 $\pm$ 0.92 & 80.04 $\pm$ 0.99\\
    OE~\cite{hendrycks2018deep} & 81.60 $\pm$ 0.35 & 56.71 $\pm$ 0.80 & 78.46 $\pm$ 0.37 & 63.59 $\pm$ 0.77 & 72.70 $\pm$ 0.41 & 71.95 $\pm$ 0.70 & 64.55 $\pm$ 0.45 & 79.39 $\pm$ 0.60\\
    OEC~\cite{wang2020few}  & 81.05 $\pm$ 0.74 & 57.68 $\pm$ 0.83 & 76.19 $\pm$ 0.51 & 64.17 $\pm$ 0.94 & 71.11 $\pm$ 0.78 &  74.46 $\pm$ 1.05 & 61.97 $\pm$ 0.86 & 80.09 $\pm$ 1.44\\
    \arrayrulecolor{gray}\cmidrule{1-9}\arrayrulecolor{black}
    ParamMix & 81.02 $\pm$ 0.36 & 55.28 $\pm$ 0.78 & 78.74 $\pm$ 0.37 & 60.16 $\pm$ 0.78 & 73.80 $\pm$ 0.40 & 67.94 $\pm$ 0.73 & 65.10 $\pm$ 0.44 & 77.32 $\pm$ 0.61\\ 
    HyperMix & \textbf{81.94 $\pm$ 0.34} & \textbf{53.67 $\pm$ 0.79} & \textbf{79.69 $\pm$ 0.36} &  \textbf{58.80 $\pm$ 0.80} & \textbf{74.87 $\pm$ 0.40} &  \textbf{67.00 $\pm$ 0.76} & \textbf{66.36 $\pm$ 0.45} &  \textbf{76.46 $\pm$ 0.64} \\
    \midrule
    & \multicolumn{8}{c}{MiniImageNet\cite{vinyals2016Matching}} \\ \midrule
    MSP~\cite{hendrycks17baseline} & 73.75 $\pm$ 0.39 & 68.16 $\pm$ 0.74 & 71.04 $\pm$ 0.41 & 71.97 $\pm$ 0.71 & 66.77 $\pm$ 0.43 & 77.01 $\pm$ 0.65 & 60.13 $\pm$ 0.45 & 82.80 $\pm$ 0.56\\
    Entropy~\cite{mukhoti2021deterministic} & 63.03 $\pm$ 0.42 & 79.87 $\pm$ 0.57 & 62.12 $\pm$ 0.41 & 81.38 $\pm$ 0.56 & 60.32 $\pm$ 0.42 & 83.28 $\pm$ 0.53 & 57.38 $\pm$ 0.44 & 85.58 $\pm$ 0.50\\
    ODIN~\cite{liang2018enhancing} & 73.99 $\pm$ 0.39 & 67.80 $\pm$ 0.74 & 71.23 $\pm$ 0.41 & 71.62 $\pm$ 0.71 & 66.87 $\pm$ 0.43 & 76.81 $\pm$ 0.65 & 60.19 $\pm$ 0.45 & 82.78 $\pm$ 0.56\\
    DM~\cite{lee2018simple} & 60.40 $\pm$ 0.18 & 84.69 $\pm$ 0.21 & 57.71 $\pm$ 0.17 & 86.61 $\pm$ 0.20 & 55.41 $\pm$ 0.17 & 87.80 $\pm$ 0.18 & 53.32 $\pm$ 0.17 & 88.77 $\pm$ 0.17\\
    pNML~\cite{bibas2021single} & 59.78 $\pm$ 0.43 & 84.03 $\pm$ 0.47 &  58.60 $\pm$ 0.49 & 85.08 +- 0.59 & 56.79 $\pm$ 0.89 & 86.31 $\pm$ 0.71 & 53.43 $\pm$ 0.98 & 87.21 $\pm$ 0.85\\
    OE~\cite{hendrycks2018deep} & 73.09 $\pm$ 0.40 & 67.90 $\pm$ 0.73 & 70.30 $\pm$ 0.42 & 72.31 $\pm$ 0.70 & 65.87 $\pm$ 0.44 & 77.39 $\pm$ 0.64 & 59.64 $\pm$ 0.47 & 82.69 $\pm$ 0.56 \\
    OEC~\cite{wang2020few}  & 73.61 $\pm$ 0.52 & 68.08 $\pm$ 0.79 & 69.94 $\pm$ 0.53 & 71.59 $\pm$ 0.84 & 65.46 $\pm$ 0.71 & 78.46 $\pm$ 0.85 & 60.22 $\pm$ 0.76 & 82.17 $\pm$ 0.81\\
    \arrayrulecolor{gray}\cmidrule{1-9}\arrayrulecolor{black}
    ParamMix & 73.46 $\pm$ 0.41 & 67.47 $\pm$ 0.73 & 70.88 $\pm$ 0.43 & 71.17 $\pm$ 0.69 & 66.03 $\pm$ 0.48 & 76.85 $\pm$ 0.63 & 59.83 $\pm$ 0.43 & 81.80 $\pm$ 0.59\\ 
    HyperMix & \textbf{76.02 $\pm$  0.37} & \textbf{65.22 $\pm$ 0.73} & \textbf{74.09 $\pm$ 0.38} & \textbf{70.61 $\pm$ 0.70} & \textbf{69.67 $\pm$ 0.40} & \textbf{75.03 $\pm$ 0.63} & \textbf{62.55 $\pm$ 0.43} & \textbf{81.80 $\pm$ 0.55} \\
    \bottomrule
    \end{tabular}
}
\label{tab:noisy_results}%
\vspace{-4mm}
\end{table*}%

\minisection{Performance of Popular OOD Detection Methods} 
We observe that several recent OOD detection methods underperform the simple baseline MSP in the FS-OOD setting. In particular, we found that popular OOD methods such as DM and pNML had limited success in detecting OOD samples; DM for example has almost $15\%$ worse AUROC and almost $23\%$ worse FPR@90 compared to MSP, which stands in sharp contrast to its large improvements in the many-shot setting.
We posit that this is due to the limited number of support samples per episode, leading to the empirical covariance/data matrix used in these approaches to be degenerate. 
We indeed find this to be the case when comparing the singular values of the covariance matrix used in DM for few-shot and many-shot settings (see Appendix~\ref{apx:sing_vals}).
This limits the generalization ability of these methods in differentiating IND from OOD samples.

\minisection{HyperMix} 
In contrast, we observe that our proposed HyperMix performs the best. Using HyperMix at meta-training, the model not only does better on the OOD detection task, but also leads to improved IND accuracy compared to the baselines with hypernetworks. We see that the combination of both ParamMix and OOE-Mix is important. While each individually is comparable or does better than the baselines, the combination of the two significantly outperforms the baseline methods.

Why is the combination of ParamMix and OOE-Mix especially effective?
We hypothesize that each makes HyperMix robust to certain types of outliers: ParamMix smooths the inter-support class decision boundary, while OOE-Mix regularizes behavior on outside classes.
We verify this by testing ParamMix, OOE-Mix, and HyperMix on outliers created by mixing inlier class samples (INE-INE) and inliers with OOE samples (INE-OOE)  in Table~\ref{tab:mixup_ablation}; we indeed observe ParamMix and OOE-Mix have different tendencies, and that HyperMix tends to combine their strengths.

\vspace{-1mm}
\subsection{Noisy Labels}
\label{sec:noisy}
\vspace{-2mm}

\begin{figure}
    \centering
    \includegraphics[width=0.47\columnwidth]{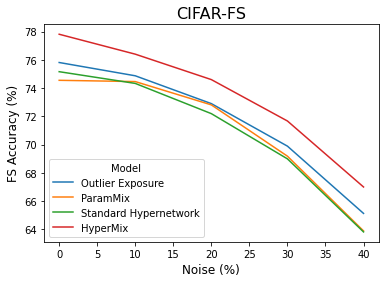}
    \includegraphics[width=0.47\columnwidth]{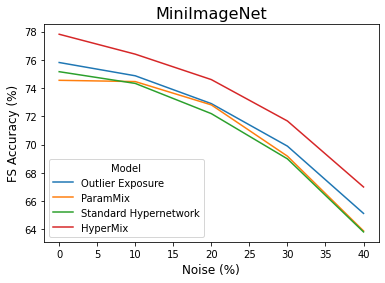}
    \vspace{-2mm}
    \caption{FS accuracy on FS-CIFAR-100 (left) and MiniImageNet (right) with varying levels of label noise in the support set.}
    \label{fig:fs_acc_noise}
    \vspace{-4mm}
\end{figure}

Even in few-shot settings, outliers can have noisy labels and could be present in the support sets, and such mislabeled samples can have a significant impact on performance~\cite{liang2022few}.
We thus also evaluate all the models in the FS-OOD settings when noisy labels are present during meta-testing, reporting OOD detection results in Table~\ref{tab:noisy_results} and few-shot accuracy in Figure~\ref{fig:fs_acc_noise}.
We restrict our experiments to hypernetwork-based few-shot learning, as it saw the strongest OOD detection performance in Table~\ref{tab:fs_ood}.
We experiment with varying levels of noise present in the support set ranging from 10\% to to 40\% of the total support samples.
For each evaluation scenario, we randomly select IND samples ($\{x, y\}$) from the support set and replace them with OOD samples $(\{\hat{x}, y\})$. 
To make this scenario especially challenging, we pair each IND class with an OOE class, to model real-world classes being similar~\cite{han2018co}.
As expected, we observe that all methods, including ours, see worse performance with higher noise levels.
On the other hand, we observe that HyperMix maintains a performance improvement gap even up to $40\%$ support set noise. 

\vspace{-2mm}
\section{Conclusion}
\vspace{-2mm}
Standard few-shot methods are unequipped to deal with OOD samples during test time and will tend to misclassify them as one of the IND classes by default.
At the same time, while recently proposed methods have led to improvements in OOD detection in settings with abundant in-distribution data, we find that many of them struggle in few-shot settings, failing to outperform the simple MSP~\cite{hendrycks17baseline} baseline. 
To address this gap, we show that a hypernetwork-based framework for FS-OOD outperforms ProtoNet or fine-tuning based approaches.
We further show that exposing the model to out-of-episode (OOE) outliers mixed with the query set during meta-training helps prepare the model for test OOD samples, and mixup of hypernetwork weights generated from the support set further helps the model generalize.
This combination of augmentation techniques, which we call HyperMix, strongly outperforms baseline methods in FS-OOD experiments on FS-CIFAR-100 and MiniImageNet in a variety of settings.

{\small
\bibliographystyle{ieee_fullname}
\bibliography{refs}
}

\begin{acronym}[CCCCCCCC]
    \acro{OOD}{Out-of-Distribution}
    \acro{IND}{In-Distribution}
    \acro{FSC}{Few-Shot Classification}
\end{acronym}

\begingroup
\clearpage
\appendix

\section{Baseline adaptations for OOD detection in the few-shot classification setting}
\label{apx:baseline_adaptations}
As reported in Section~\ref{sec:baselines}, we adapt several existing OOD detection methods for the few-shot classification setting. In particular, the adaptations are carried out for the hypernetwork model $\mathcal{M}$ learned for few-shot classification (FSC). The OOD detection metrics (AUROC and FPR@90) reported in Section~\ref{sec:results} are calculated using the IND score, which we denote as $s_{\mathcal{M}}$. For all baselines, we use the same implementations for the feature extractor $\mathcal{F}$ and the hypernetwork $\mathcal{H}$ as described in Section~\ref{sec:setup}.

Below we provide details on how $s_{\mathcal{M}}$ is calculated for each method.
\begin{enumerate}
    \item MSP~\cite{hendrycks17baseline}: The MSP uses the largest value of the softmax as the IND score, \ie,
    \begin{align}
        s_\mathcal{M}(x_q; \mathcal{D}_s) = \max_{c \in C_n}
        p(Y = c|x_q)
    \end{align}
    where $p(Y = c|x_q) = \text{softmax}\left(\mathcal{C}\left(\mathcal{F} \left(x_q\right)|W\right)\right)[c]$, and $C_n$ corresponds to the set of labels in a few-shot episode as described in Section~$\ref{sec:background}$.
    \item Entropy~\cite{mukhoti2021deterministic}: We also consider using the Shannon entropy that depicts the uncertainty with respect to the probabilities predicted by the model. Specifically, we compute 
    \begin{align}
        s_\mathcal{M}(x_q; \mathcal{D}_s) &:= -H(Y) \\ &= \sum_{c \in C_n} p(Y = c|x_q) \cdot \log\left(p(Y = c|x_q)\right),
    \end{align}
    where $p(Y = c|x_q) = \text{softmax}\left(\mathcal{C}\left(\mathcal{F} \left(x_q\right)|W\right)\right)[c]$.
    \item ODIN~\cite{liang2018enhancing}: The ODIN detector uses temperature scaling and small input perturbation to improve the MSP baseline for OOD detection. In particular, ODIN introduces hyperparameter $S$ for temperature scaling and $\epsilon$ for the magnitude of input perturbation that are used to adjust the IND confidence score:
    \begin{align}
        s_\mathcal{M}(x_q; \mathcal{D}_s, T, \epsilon) = \max_{c \in C_n} p(\tilde{Y} = c|\tilde{x}_q; T)
    \end{align}
    where $\tilde{x}_q$ is the perturbed input calculated as $$\tilde{x}_q = x_q - \epsilon \cdot \text{sign} \left(-\nabla_{x_q} \log\left( \max_{c \in C_n} p(Y = c|x_q; T) \right)\right),$$ $p(Y|x_q; T)$ is the softmax probability with temperature scaling, and $p(\tilde{Y}|\tilde{x}_q; T)$ denotes the temperature scaled probability of the perturbed input. In our hypernetwork framework, we calculate $p(Y = c|x_q; T) = \text{softmax}\left(\mathcal{C}\left(\mathcal{F} \left(x_q\right)|W\right)\right)[c]$.
    \item DM~\cite{lee2018simple}: The DM OOD detector uses the features extracted from each block of the encoder $\mathcal{F}$. We denote the feature extracted at each block as $f(x,l)$. Given the labels in the support set along with their corresponding features, the parameters of the Gaussian distribution $\mu_{l,c}$ and $\Sigma_{l}$ is fitted to each block for each class $c$ in the episode. The covariance is shared across all the classes. The layer-specific score $s_l$ for query $x_q$ is computed as:
    \begin{align}
        s_l(x_q; \mathcal{D}_s) &:= \max_{c \in C_n} \log \left( \mathcal{N}\left(f\left(x_q, l\right);\mu_{l,c}, \Sigma_l\right) \right)
    \end{align}
    The final score can be computed using a linear combination of layer-specific scores, \ie $s_\mathcal{M}(x_q; \mathcal{D}_s) := \alpha_l \cdot s_l(x_q; \mathcal{D}_s)$, where the logistic regression model with parameters $\alpha_l$ are found using the validation dataset. However, we found that learning the logistic regression model overfits the support set containing few samples and the OOD samples used in the validation set. Instead, we use $s_\mathcal{M}(x_q; \mathcal{D}_s) := \max_l s_l(x_q; \mathcal{D}_s)$ to work better and we use that in all our experiments. Note that DM doesn't use the adapted task-specific classifier to detect OOD samples.
    
    \item pNML~\cite{bibas2021single}: Predictive Normalized Maximum Likelihood (pNML) learner is a recently proposed OOD detection approach that uses generalization error to detect OOD samples. \cite{bibas2021single} derived a pNML regret for a single layer NN that can be used for features extracted from a pre-trained deep encoder, and can be used as the confidence measure for detecting OOD samples. Namely, the pNML regret of a single layer NN is
    \begin{align}
        \Gamma(x; \mathcal{D}_s) = \log \sum_{i=1}^{C_n} \frac{p_i}{p_i+{p_i}^{x^Tg}(1-p_i)}
    \end{align}
    where $p_i$ is the output probability for the $i^{th}$ class and $g$ is a function of the inverse of the data matrix of the features of the IND data as defined in Equation 12 of \cite{bibas2021single}. We defined the IND score as $s_\mathcal{M}(x_q; \mathcal{D}_s) := 1 - \Gamma(x; \mathcal{D}_s)$. In our hypernetwork-based framework for few-shot classification, we use the output probabilities from the adapted task-specific classifier to define $p_i$. The pNML parameter $g$ is calculated using the features in the support set of a given episode during meta-testing. Due to the limited number of IND samples in the support set, we found the pNML regret to have limited success in detecting OOD samples in the few-shot setting.
    
    \begin{table}[!t]
\centering
\caption{The candidate values and the final determined values for the hyperparameters of all the methods.}
\resizebox{\columnwidth}{!}{
\begin{tabular}{l||lcc}
\toprule
\multicolumn{1}{l}{Method} & \multicolumn{1}{l}{Candidates} & \multicolumn{2}{c}{Determined} \\
\midrule
& & \multicolumn{1}{c}{5-shot 5-way} & \multicolumn{1}{c}{10-shot 5-way} \\
\midrule
& \multicolumn{3}{c}{FS-CIFAR-100~\cite{bertinetto2019meta}} \\ \midrule
ODIN~\cite{liang2018enhancing} & $\epsilon \in \{0.2, 0.02, 0.002\}$ & $\epsilon=0.002$ & $\epsilon=0.002$ \\
& $T \in \{0.1, 1.0, 10.0\}$ & $T=1.0$ & $T=10.0$ \\
\arrayrulecolor{gray}\cmidrule{1-4}\arrayrulecolor{black}
OE~\cite{hendrycks2018deep} & $\beta \in \{0.1, 1.0, 10.0\}$ & $\beta=1.0$ & $\beta=1.0$\\
\arrayrulecolor{gray}\cmidrule{1-4}\arrayrulecolor{black}
ParamMix & $a_{PM} \in \{0.1, 1.0, 2.0\}$ & $a_{PM} = 2.0$ & $a_{PM} = 2.0$\\
& $b_{PM} \in \{5.0, 10.0, 20.0\}$ & $b_{PM} = 5.0$& $b_{PM} = 5.0$\\
\arrayrulecolor{gray}\cmidrule{1-4}\arrayrulecolor{black}
OOE-Mix  & $a_{OM} \in \{1.0, 10.0, 20.0\}$ & $a_{OM} = 20.0$ & $a_{OM} = 20.0$\\
& $b_{OM} \in \{1.0, 10.0, 20.0\}$ & $b_{OM} = 20.0$ & $b_{OM} = 20.0$\\
\midrule
& \multicolumn{3}{c}{MiniImageNet~\cite{vinyals2016Matching}} \\ \midrule
ODIN~\cite{liang2018enhancing} & $\epsilon \in \{0.2, 0.02, 0.002\}$ & $\epsilon=0.002$ & $\epsilon=0.002$ \\
& $T \in \{0.1, 1.0, 10.0\}$ & $T=1.0$ & $T=10.0$ \\
\arrayrulecolor{gray}\cmidrule{1-4}\arrayrulecolor{black}
OE~\cite{hendrycks2018deep} & $\beta \in \{0.1, 1.0, 10.0\}$ & $\beta=1.0$ & $\beta=1.0$\\
\arrayrulecolor{gray}\cmidrule{1-4}\arrayrulecolor{black}
ParamMix & $a_{PM} \in \{0.1, 1.0, 2.0\}$ & $a_{PM} = 0.1$ & $a_{PM} = 0.1$ \\
& $b_{PM} \in \{5.0, 10.0, 20.0\}$ & $b_{PM} = 10.0$ & $b_{PM} = 10.0$\\
\arrayrulecolor{gray}\cmidrule{1-4}\arrayrulecolor{black}
OOE-Mix  & $a_{OM} \in \{1.0, 10.0, 20.0\}$ & $a_{OM} = 10.0$ & $a_{OM} = 10.0$\\
& $b_{OM} \in \{1.0, 10.0, 20.0\}$ & $b_{OM} = 10.0$ & $b_{OM} = 10.0$\\
\bottomrule
\end{tabular}
}
\label{tab:hyperparams}%
\end{table}%
    
    \item OE~\cite{hendrycks2018deep}: The Outlier Exposure (OE) uses additional OOE samples during meta-training stage to learn the model. To use OE, we apply the following training objective when meta-training the hypernetwork framework:
    \begin{equation}
    \begin{split}
        \mathcal{L}_{OE} = ~&\Ebb_{\{x,y\} \sim \mathcal{D}_q^{IND}} \left[\mathcal{L}_{CCE}\left(\mathcal{M}\left(x_q; \mathcal{D}_s\right), y \right)\right] + \\ & \beta \cdot \Ebb_{\left( \tilde{x}, \tilde{y} \right) \sim \mathcal{D}_q^{OOE}} \left[\mathcal{L}_{CCE} \left(\mathcal{M}\left(\tilde{x}_q; \mathcal{D}_s\right), \tilde{y}\right)\right]
    \end{split}        
    \end{equation}
    where $\mathcal{M}(x_q; \mathcal{D}_s)$ are the output logits for query $x_q$ in the episode containing the support set $\mathcal{D}_s$, $\mathcal{L}_{CCE}$ computes the cross entropy loss, $\beta$ is a hyperparameter that weighs the loss for OOE samples in the episode, and $\mathcal{D}_q^{IND}$ and $\mathcal{D}_q^{OOE}$ are the IND and OOE samples in the queries of a given episode.
    \item OEC~\cite{wang2020few}: Similar to OE, Out-of-Episode Classifier (OEC) leverages OOE inputs at the meta-training stage. However, unlike OE which uses uniform label distribution for a multi-class objective, OEC learns a binary classifier that distinguishes IND classes from OOE classes. The binary classifier uses the following objective during meta-training stage
    \begin{equation}
    \begin{split}
        \mathcal{L}_{OEC} = ~& -\sum_{x_q \in \mathcal{D}_q^{IND}} \log \left(\sigma\left( s\left(x_q; \mathcal{D}_s\right) \right)\right) - \\ &\sum_{\tilde{x}_q \in \mathcal{D}_q^{OOE}} \log \left(1 - \sigma\left( s\left(\tilde{x}_q; \mathcal{D}_s\right) \right)\right)
    \end{split}
    \end{equation}
    where $s_\mathcal{M}(x_q; \mathcal{D}_s) = \max_{c \in C_n} \log p(Y = c|x_q)$, and $\mathcal{D}_q^{IND}$ and $\mathcal{D}_q^{OOE}$ are the IND and OOE samples in the queries of a given episode. In the few-shot setting, we use the prediction as $\hat{y} = \text{argmax}_{c \in C_n} \log p(Y = c|x_q)$ during meta-testing for calculating the accuracy.

\end{enumerate}

\section{Hyperparameters}
\label{apx:hyperparameters}
The hyperparameters used for the various baseline methods in our experiments are included in Table~\ref{tab:hyperparams}, showing the settings tried and the value providing the best results. For OOE-Mix and ParamMix, we use a random draw from the Beta distribution to determine the mixup coefficient $\lambda$ per sample, as is common practice for Mixup~\cite{zhang2018mixup}. We use MSP~\cite{hendrycks17baseline} for detecting OOD samples after meta-training, using the proposed HyperMix approach. We find the best hyperparameter settings to be fairly consistent across number of shots and the dataset.
\endgroup

\begin{figure}
\centering
\vspace{-1.5em}
\includegraphics[width=0.45\textwidth]{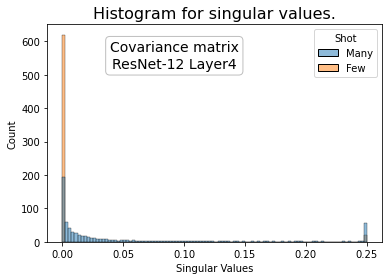}
\vspace{-0.5em}
\caption{Covariance matrix singular values distribution for many- and few-shot. In the few-shot case, the empirical covariance matrix is degenerate, which leads OOD methods like DM~\cite{lee2018simple} to fail.}
\label{fig:rank_cov}
\vspace{-3mm}
\end{figure}

\section{Effects of few samples in OOD detection}
\label{apx:sing_vals}
Many out-of-distribution detection methods operate by learning the statistics of the in-distribution samples in some manner or another. 
In the typical OOD setting, it is common for there to be thousands of examples per class for a model to learn a strong understanding of the in-distribution classes.
In contrast, in our FS-OOD, the model has very limited examples in the support set, which may conflict with assumptions made by some OOD methods.
An example of this is Deep Mahalanobis~\cite{lee2018simple}, which estimates the class-conditional mean and covariance per class.
With a few examples, it is difficult to get a good estimate on such statistics. 
As shown in Figure~\ref{fig:rank_cov}, the data matrix for few examples is degenerate in few-shot cases, leading to especially bad performance (Table~\ref{tab:fs_ood}).

\end{document}